%% file: main.tex
\documentclass[11pt]{article}

\usepackage[T1]{fontenc}
\usepackage[utf8]{inputenc}
\usepackage{lmodern}
\usepackage{textcomp}      
\usepackage{microtype}     
\usepackage{float}
\usepackage[margin=1in]{geometry}
\usepackage{booktabs,tabularx,adjustbox}
\usepackage{adjustbox}
\usepackage{graphicx,caption,subcaption}
\usepackage[numbers]{natbib} 
\usepackage{booktabs}
\usepackage{float}
\usepackage{chngcntr}
\usepackage{booktabs,tabularx,makecell,threeparttable,caption}
\usepackage{longtable,booktabs,makecell}
\usepackage{rotating} 
\usepackage{booktabs}
\usepackage{array}
\usepackage{makecell}
\usepackage{graphicx}
\graphicspath{{figures/}}
\DeclareGraphicsExtensions{.pdf,.png,.jpg}
\usepackage{url}
\usepackage[hidelinks]{hyperref}
\usepackage[section]{placeins}
\usepackage{graphicx,caption,subcaption}
\usepackage{float}
\usepackage{placeins} 
\usepackage{adjustbox}
\usepackage{booktabs}
\usepackage{tabularx}
\usepackage{rotating} 
\usepackage{array} 
\usepackage{makecell}

\usepackage{amsmath,amssymb,bm}
\usepackage{graphicx}
\usepackage{subcaption}     
\usepackage{booktabs,tabularx,makecell,array}
\usepackage[hidelinks]{hyperref}
\usepackage{fontawesome5} 
\usepackage{xcolor}        
\usepackage{fancyhdr}      
\usepackage{hologo}       
\definecolor{githubcolor}{RGB}{0, 0, 0} 
\newcommand{\github}{\href{https://github.com/Abitsfhuusrtyt/-Geometric-Mixture-Classifier-GMC---A-Discriminative-Per-Class-Mixture-of-Hyperplanes}{\textcolor{githubcolor}{\faIcon[brands]{github}}}}
\newcolumntype{C}{>{\centering\arraybackslash}X}
\newcolumntype{L}{>{\raggedright\arraybackslash}X}

\usepackage{authblk}

\title{Geometric Mixture Classifier (GMC)\\
\large A Discriminative Per-Class Mixture of Hyperplanes for Fast, Transparent Classification}

\author[1]{Prasanth K. K.}
\author[2]{Shubham Sharma}

\affil[1]{Ooty, The Nilgiris, Tamil Nadu, India - 643005 \\ 
\texttt{abiprasanth0101@gmail.com}}

\affil[2]{Founder \& CEO, SunitechAI \\ 
\url{www.sunitechai.com} \\ 
\texttt{shubham.sharma@sunitechai.com}}

\date{}

\begin{document}
\maketitle

\fancypagestyle{firstpage}{
  \fancyhf{} 
  \renewcommand{\headrulewidth}{0pt} 
    \fancyfoot[C]{
    \small \github~Code and data: \url{https://github.com/Abitsfhuusrtyt/-Geometric-Mixture-Classifier-GMC---A-Discriminative-Per-Class-Mixture-of-Hyperplanes}
  }
}
\thispagestyle{firstpage}
\begin{abstract}
Many real world categories are multimodal: a single class occupies several disjoint regions of feature space. Classical linear models (e.g., logistic regression, linear SVM) impose one global hyperplane and therefore struggle on such data, while high capacity alternatives (kernel SVMs, deep nets) can fit the structure but often sacrifice interpretability, require heavier tuning, and incur higher computational cost.

We propose the Geometric Mixture Classifier (GMC), a discriminative model that represents each class as a mixture of hyperplanes. Within a class, GMC aggregates plane scores with a temperature controlled soft-OR (log sum exp) that smoothly approximates the maximum; across classes, it applies a standard softmax to produce probabilistic posteriors. GMC supports an optional Random Fourier Features (RFF) mapping that equips it with nonlinear power while keeping inference linear in the number of planes and lifted features.

A practical training recipe : geometry aware initialization via k-means, automatic plane budgeting via silhouette score, alpha annealing, usage aware $L_2$ regularization, label smoothing, and early stopping, makes GMC plug and play in practice. Experiments on synthetic multimodal benchmarks (moons, circles, anisotropic blobs, two spirals) and standard tabular/image features (iris, wine, WDBC breast cancer, digits) show that GMC consistently outperforms linear baselines and k-NN, and is competitive with RBF-SVM, Random Forests, and compact MLPs, and yields transparent geometric introspection via plane and class level responsibility visualizations.

Because inference scales linearly in the number of planes and feature dimension, GMC is CPU friendly; on average, inference takes single-digit microseconds per example, typically faster than or comparable to RBF-SVM and compact MLPs. With post hoc temperature scaling, ECE drops from $\approx0.06$ to $\approx0.02$ on average. GMC therefore occupies a favorable point on the accuracy interpretability efficiency spectrum: more expressive than linear models, lighter and more transparent than kernel or deep models.
\end{abstract}

\section{Introduction}

Supervised classification is a central problem in machine learning and data science.
From medical diagnosis and fraud detection to document tagging and vision, practitioners routinely face datasets where classes are heterogeneous and \emph{multimodal}, that is, a single semantic class occupies several disjoint regions of feature space.
Classical linear models such as logistic regression and linear SVMs are appealing for their convex training, simplicity, and interpretability, but their single global hyperplane struggles on multimodal structures.
Kernel machines can capture nonlinearity, yet they often entail higher computational cost, memory footprint, and less transparent decision rules.
Deep neural networks excel in accuracy but can require significant tuning, large compute budgets, and typically provide limited geometric introspection into how decisions are made.

\paragraph{Problem definition.}
We seek a classifier that (i) models multimodal class structure without resorting to heavy black box architectures; (ii) remains data and compute efficient; and (iii) provides geometric interpretability, clarifying how, where, and by which sub model a sample is classified. Existing approaches typically trade off at least one of these considerations: linear models lack expressivity; kernel methods and ensembles add complexity and cost; and deep models, while powerful, obscure local decision geometry.

\paragraph{Our contributions.}
We address this gap with the \emph{Geometric Mixture Classifier (GMC)}, a per class mixture of hyperplanes combined by a soft-OR operator. Our main contributions are:
\begin{itemize}
  \item \textbf{Model.} GMC represents each class $c$ with $M_c$ linear experts $(\mathbf{w}_{c,m}, b_{c,m})$. Within a class, scores are pooled via a temperature-controlled log sum exp (soft-OR) that smoothly approximates the maximum; across classes, a standard softmax produces probabilistic posteriors.
  \item \textbf{Nonlinear extension.} GMC supports an optional Random Fourier Features (RFF) mapping that equips it with nonlinear power while preserving linear time inference in the lifted feature dimension. An \texttt{auto} mode chooses between linear and RFF based on validation likelihood, ensuring robustness across linear and nonlinear regimes.
  \item \textbf{Training recipe.} A practical pipeline geometry aware initialization via k-means, automatic plane budgeting via silhouette score, $\alpha$ annealing (soft $\rightarrow$ sharp pooling), label smoothing, early stopping, and usage aware $\ell_2$ regularization—makes GMC stable to train with minimal hyperparameter tuning.
  \item \textbf{Interpretability.} GMC exposes per class plane responsibilities and a soft union of half spaces geometry in the working feature space. We provide diagnostics and visualizations (decision regions, plane overlays, responsibility maps, reliability diagrams) that make the model’s decisions transparent.
  \item \textbf{Empirical evaluation.} Across synthetic multimodal datasets (moons, circles, anisotropic blobs, two spirals) and standard benchmarks (iris, wine, WDBC breast cancer, digits), GMC consistently outperforms linear baselines and $k$-NN, is competitive with RBF-SVM, Random Forests, and compact MLPs, and achieves low latency CPU inference. With post hoc temperature scaling, calibration improves substantially (ECE $\approx 0.06 \rightarrow 0.02$ on average).
\end{itemize}

\noindent\textbf{In summary, }
GMC directly encodes the idea that a class is a soft union of locally linear regions, augments it with usage aware regularization for reliability, and demonstrates strong accuracy, interpretability, and efficiency with a light computational footprint.

\section{Related Work}

\paragraph{Linear models.}
Logistic regression and linear SVMs are strong baselines thanks to convex training, efficiency, and interpretability.
Their inductive bias one global hyperplane per class or margin works well when each class is roughly unimodal, but degrades when class regions are scattered or disjoint.
GMC preserves the linear flavor but allocates multiple local hyperplanes per class, directly addressing multimodality.

\paragraph{Kernel methods.}
Kernel SVMs (especially with RBF kernels) can model complex, nonlinear structure and often achieve excellent accuracy on small to medium sized datasets.
However, their decision geometry is implicit in the kernel expansion, inference cost grows with the number of support vectors, and interpretability is limited.
GMC attains similar flexibility via an explicit union of half spaces (optionally in an RFF lifted feature space) and maintains explicit, interpretable linear components with predictable inference cost.

\paragraph{Ensemble methods.}
Tree ensembles such as Random Forests and Gradient Boosted Trees (e.g., XGBoost, LightGBM) are highly effective at capturing heterogeneous decision boundaries.
They offer feature importance summaries and partial dependence or SHAP plots, but the overall decision surface is a large ensemble of trees that is less geometrically transparent.
GMC, by contrast, provides explicit, piecewise linear acceptance regions per class with clear geometric meaning.

\paragraph{Generative mixture models.}
Gaussian mixture models represent classes via density estimation and classify by comparing class conditional likelihoods.
They capture multimodality but rely on distributional assumptions and optimize likelihood rather than direct classification loss.
GMC is discriminative: it models separating geometry rather than densities and produces probabilistic posteriors via a class level softmax, with post hoc temperature scaling used when improved calibration is desired.

\paragraph{Mixture of experts (MoE).}
Classical MoE frameworks train both experts and a separate gating network that assigns input dependent weights, combining expert outputs by weighted averaging.
GMC shares the ``many local experts'' idea but differs in key ways:
(i) there is no separate gating network responsibilities emerge from log sum exp pooling;
(ii) experts are organized per class and combined by a soft-OR (approximate max), not by averaging; and
(iii) Usage aware regularization discourages permanently idle experts.

\paragraph{Maxout and latent part models.}
Maxout networks pool linear units with a max operator, and latent part models in vision use a max over components.
GMC adopts a similar principle but applies it within each class, using a tunable soft-OR (log sum exp at temperature $\alpha$) for smoother optimization and an explicit class-level softmax for probabilistic outputs.

\paragraph{Prototype methods and nearest neighbors.}
Prototype-based classifiers (e.g., learning vector quantization) and nearest neighbor methods classify points by similarity to stored exemplars.
Although simple and often competitive, they can yield irregular decision boundaries and degrade in high dimensions without careful metric learning.
GMC yields globally consistent, smooth piecewise linear boundaries while still exposing local responsibilities that provide prototype-like interpretability.

\paragraph{Naive Bayes and LDA.}
Naive Bayes and Linear Discriminant Analysis are classical generative classifiers that yield linear decision functions under standard assumptions (e.g., Gaussian NB with shared variance, LDA with shared covariance).
GMC is discriminative, does not assume feature independence or Gaussianity, and instead allocates capacity to multiple separating hyperplanes per class.

\paragraph{Interpretable machine learning.}
Post hoc explanation methods such as LIME and SHAP approximate local decision behavior for otherwise black-box models.
Although useful, they provide surrogates rather than intrinsic explanations.
GMC is intrinsically interpretable: predictions can be explained directly in terms of which class level score dominated and which plane(s) within that class carried the responsibility.
\input{conceptual_table.tex}

\section{The Geometric Mixture Classifier (GMC)}

\subsection{Model formulation}

We consider multiclass classification with training data
$\{(x_i, y_i)\}_{i=1}^N$, where $x_i \in \mathbb{R}^d$ and
$y_i \in \{1,\dots,C\}$.
To capture multimodal class geometry, GMC allocates multiple hyperplanes per class rather than a single global separator.
For a unified presentation (linear or lifted), let $\phi(x)$ denote the working feature map; in the linear case $\phi(x)=x$.

\paragraph{Per plane score.}
Each plane $m$ of class $c$ defines an affine score
\begin{equation}
z_{c,m}(x) = w_{c,m}^\top \phi(x) + b_{c,m}.
\end{equation}

\paragraph{Soft-OR within a class.}
The class score is a smooth maximum of its planes:
\begin{equation}
s_c(x) = \frac{1}{\alpha} \log \sum_{m=1}^{M_c} \exp(\alpha \, z_{c,m}(x)),
\end{equation}
with temperature $\alpha > 0$. As $\alpha \to \infty$ this approaches
$\max_m z_{c,m}(x)$. The plane responsibilities (implicit gating weights) are
\begin{equation}
a_{c,m}(x) = \frac{\exp(\alpha z_{c,m}(x))}{\sum_j \exp(\alpha z_{c,j}(x))},
\end{equation}
so that $\sum_m a_{c,m}(x) = 1$.

\paragraph{Across class softmax.}
Final class probabilities are
\begin{equation}
p_c(x) = \frac{\exp(s_c(x))}{\sum_k \exp(s_k(x))}.
\end{equation}

\begin{figure}[t]

  \centering
  \includegraphics[width=1\linewidth]{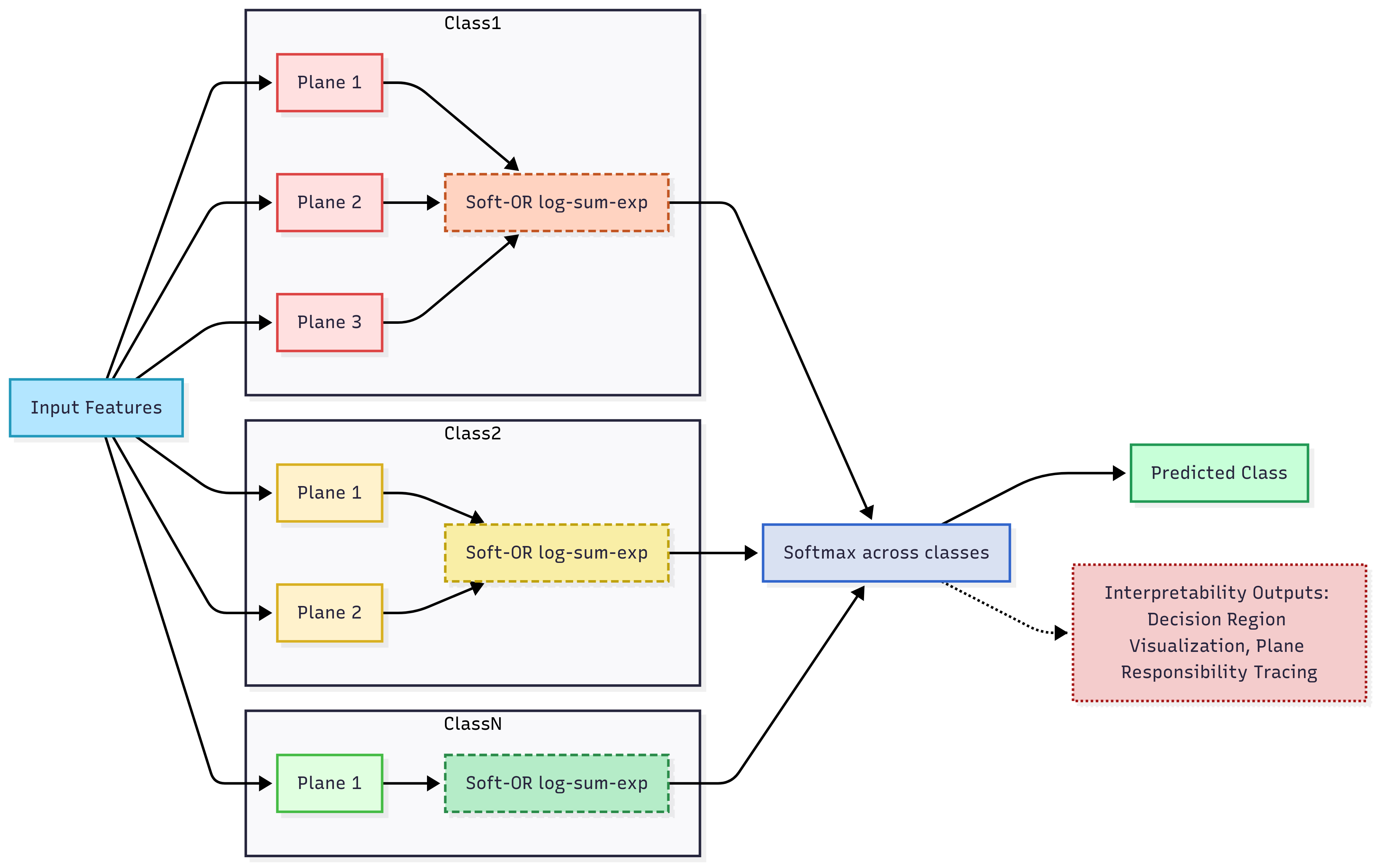}
  \caption{Model Formulation}
  \label{fig:gmc-algorithm}
\end{figure}

GMC is thus a mixture of hyperplanes per class, using soft-OR pooling within classes and softmax competition across classes as can be seen in Fig.~\ref{fig:gmc-algorithm}. When $M_c=1$ for all $c$, GMC reduces to multinomial logistic regression.

\subsection{Assumptions, Approximation Capacity, and Complexity}
\label{sec:assumptions_capacity}
\paragraph{Assumptions.} GMC operates in a working feature space $\phi(x)$ (identity for linear; an RFF or other lift otherwise). For each class $c$, GMC forms a \emph{soft union of half spaces} by adding logarithmic sum exp to $M_c$ affine scores $z_{c,m}(x)=w_{c,m}^\top\phi(x)+b_{c,m}$, followed by a softmax between classes. We assume i.i.d.\ data; for auto plane budgeting we assume mild clusterability of within class structure.
\paragraph{Approximation capacity (proposition).} In the linear case ($\phi(x)=x$), the $\alpha\!\to\!\infty$ limit of GMC realizes class regions that are finite unions of half spaces (polyhedral unions), strictly generalizing multinomial logistic regression ($M_c{=}1$). Increasing $\sum_c M_c$ allows for approximation of piecewise-linear decision boundaries to arbitrary precision. With a nonlinear lift $\phi$, GMC approximates piecewise linear boundaries in the lifted space; random Fourier features provide universal approximation for shift invariant kernels.
\paragraph{Time/space complexity.} Let $d'$ be $\dim(\phi(x))$ and $M_\text{tot}:=\sum_c M_c$. Per example inference is $O(d' M_\text{tot})$ (dot products), parameters $O(d' M_\text{tot})$ plus $O(M_\text{tot})$ biases. Training per epoch cost is linear in data size and $M_\text{tot}$ with minibatch Adam and early stopping.

\subsection{Relation to mixture of experts}

GMC is MoE like but dispenses with a separate gating network.
Planes act as local experts; responsibilities $a_{c,m}(x)$ emerge directly from log sum exp pooling; and experts within a class combine by approximate maximization (a soft union of half spaces in the working feature space).
A prediction can be explained by (i) which class score won and (ii) which plane(s) in that class carried most responsibility.

\subsection{Training and regularization}

We train with cross entropy, optionally using label smoothing and class weights.
With smoothed targets
\[
y^{(\varepsilon)}_{i,c} = (1-\varepsilon)\,\mathbf{1}\{y_i=c\} + \tfrac{\varepsilon}{C},
\]
the loss is
\begin{equation}
\mathcal{L}_{\text{CE}} = -\frac{1}{N}\sum_{i,c}  \, y^{(\varepsilon)}_{i,c} \, \log p_c(x_i).
\end{equation}

\paragraph{Usage aware $L_2$.}
To discourage dead experts and promote specialization, each plane’s weight vector is penalized more when it is rarely used.
Let $u_{c,m}$ be the average responsibility of plane $m$ for class $c$ (estimated from the current minibatch or a running average).
The per plane coefficient is
\begin{equation}
\lambda_{c,m} = \lambda \left(1 + \frac{\beta}{u_{c,m} + \delta}\right),
\end{equation}
with $\beta > 0$ and a small floor $\delta > 0$.
The total objective is
\begin{equation}
\mathcal{L} = \mathcal{L}_{\text{CE}} +
\sum_{c,m} \lambda_{c,m} \, \|w_{c,m}\|_2^2,
\end{equation}
where biases are not penalized.

\paragraph{Initialization.}
We use a small set of practical inits and pick the most stable:
\begin{itemize}
  \item \textbf{k-means:} orient planes toward per class cluster centers;
  \item \textbf{logistic regression seeds:} class vs rest directions with noise;
  \item \textbf{random:} Gaussian weights.
\end{itemize}
An auto mode tries k-means and falls back if clustering is unstable.

\paragraph{Alpha annealing.}
Training starts with a small $\alpha$ (soft pooling) and increases it toward a larger target (sharper pooling), preventing early collapse to a single plane.
Early stopping, gradient clipping, and a standard optimizer (mini batch Adam with a cosine or exponential schedule) further stabilize training.

\subsection{Optimization and Convergence Considerations}
\label{sec:opt}
Training is non convex once $M_c{>}1$. We observe stable convergence due to: (i) geometry aware initialization (k-means or one vs rest logistic), (ii) $\alpha$ annealing (soft pooling $\rightarrow$ specialization $\rightarrow$ sharpening), and (iii) usage aware $\ell_2$ (stronger penalty on rarely used planes). Ablations (Appendix.~\ref{app:ablations}) show removing any of these degrades accuracy or increases variance across seeds.

\subsection{Nonlinear lifts via Random Fourier Features}

To capture smooth nonlinear boundaries while retaining linear time inference, we apply an optional RFF mapping
\begin{equation}
\phi(x) = \sqrt{\tfrac{2}{D}} \,\big[ \cos(\Omega^\top x + b); \; \sin(\Omega^\top x + b) \big],
\end{equation}
with $\Omega \sim \mathcal{N}(0, 2\gamma I)$ and $b \sim \mathrm{Unif}[0,2\pi]$.
GMC then learns mixtures of hyperplanes in $\mathbb{R}^{2D}$.
Computational cost scales as $O(D \sum_c M_c)$ for inference.
An auto mode evaluates linear vs.\ RFF variants on held out likelihood and selects the better one.

\subsection{Implementation details}

\begin{itemize}
  \item \textbf{Optimization:} mini batch Adam with cosine or exponential learning rate schedule; gradient norm clipping; early stopping on validation loss.
  \item \textbf{Preprocessing:} standardization; optional PCA (e.g., retain 95\% variance); optional RFF.
  \item \textbf{Inference cost:} $O(d' \sum_c M_c)$ dot products, where $d'$ is the post preprocessing dimension ($d'=d$ for linear GMC and $d'=2D$ with RFF).
  \item \textbf{Outputs and calibration:} GMC produces probabilistic posteriors via the across class softmax; when improved calibration is desired, we apply post hoc temperature scaling.
\end{itemize}
\subsection{Training Pipeline (Pseudo code)}
\label{sec:training_pipeline}
\begin{quote}\ttfamily\small
\textbf{Input:} $(x_i,y_i)_{i=1}^N$, classes $1{:}C$; hyperparams $M_c$ (or ``auto''), $\alpha$ schedule, $\lambda$ (L2), $\beta$ (usage penalty), label smoothing $\varepsilon$\\
\textbf{Preprocess:} standardize; optional PCA; optional RFF lift $\phi(x)$ (auto select via probe).\\
\textbf{Plane budgeting:} if auto, choose $M_c$ via silhouette on class $c$ features; cap $M_c$ if desired.\\
\textbf{Init:} (a) k-means per class $\Rightarrow$ plane seeds; else (b) one vs rest logistic replicated per class with small noise; else (c) small random.\\
\textbf{Loop (epochs):} 
\begin{enumerate}\itemsep0pt
\item For each minibatch: compute $z_{c,m}(x)$, $a_{c,m}(x)$, $s_c(x)$, $p_c(x)$.
\item Compute cross entropy (with smoothing) and usage aware $\ell_2$; backprop gradients.
\item Clip global grad norm; Adam update; optional weight norm cap.
\end{enumerate}
\textbf{Schedules:} cosine or exp LR decay; increase $\alpha$ from soft $\to$ sharp pooling.\\
\textbf{Validation:} early stop on val loss; restore best weights.\\
\textbf{Output:} calibrated probabilities (optional temperature scaling).
\end{quote}

\section{Experiments}
\label{sec:experiments}

\subsection{Experimental setup}

\paragraph{Datasets.}
We evaluate GMC on a mix of synthetic multimodal datasets (to probe geometry) and standard tabular/image benchmarks (to probe generalization, calibration, and efficiency).
\begin{itemize}
  \item Iris (UCI; 150 samples, 3 classes, 4 features).
  \item Wine (UCI; 178 samples, 3 classes, 13 features).
  \item Breast Cancer (WDBC; UCI; 569 samples, 2 classes, 30 features).
  \item Digits (scikit learn; 1{,}797 samples, 10 classes, 64 features).
  \item Moons (synthetic; 4{,}000 samples, 2 classes, 2 features; noise=0.25).
  \item Circles (synthetic; 4{,}000 samples, 2 classes, 2 features; concentric).
  \item Anisotropic blobs (synthetic; 4{,}500 samples, 3 classes, 2 features; rotated).
  \item Two spirals (synthetic; 2,000 samples, 2 classes, 2 features; highly nonlinear, noise free).
\end{itemize}

\paragraph{Baselines.}
We compare against widely used classifiers: Logistic Regression (LR), Linear SVM, RBF-SVM, Random Forest (RF), Gradient Boosted Trees (XGBoost and LightGBM), $k$ Nearest Neighbors (k-NN), and a shallow Multi Layer Perceptron (MLP).
Implementations use \texttt{scikit learn} or standard libraries.
Hyperparameters are selected on a validation split using typical grids (e.g., $C$ for SVMs, depth/trees for RF/GBM, hidden units for MLP), with early stopping where applicable.

\paragraph{Protocol.}
For each dataset we perform stratified train/validation/test splitting (validation taken from the training set).
All methods share the same splits and preprocessing pipeline.
Results are averaged over three random seeds and reported as mean~$\pm$~standard deviation.

\paragraph{Metrics.}
We report Accuracy (primary), Macro-F1 (balanced performance), Expected Calibration Error (ECE (post temperature scaling; 15 bins), Training time (seconds), and Inference time (microseconds per example, CPU).
Calibration is evaluated both before and after temperature scaling.

\paragraph{Implementation details.}
GMC defaults: automatic per class plane selection via silhouette (capped at four planes per class), $\alpha$ annealing (soft$\rightarrow$sharp pooling), usage aware $\ell_2$ regularization, early stopping (patience 12), and mini batch Adam.
For high dimensional inputs, PCA retains 95\% variance prior to any RFF mapping.
In an auto RFF mode, the method chooses between linear and RFF lifts by validation likelihood.
Timing is measured single threaded with batch size 1 (hardware details in the appendix).

\subsection{Results and analysis}

\paragraph{Main results.}
Table~\ref{tab:main-results} shows that GMC outperforms linear baselines (LR, Linear SVM) across all datasets, with the largest margins on multimodal sets (moons, circles, spirals).
On nonlinear datasets, GMC with RFF matches or surpasses RBF-SVM and MLP.
On tabular benchmarks (iris, wine, WDBC), linear GMC suffices and attains strong accuracy with minimal tuning.
On digits, GMC+RFF is competitive with RBF-SVM and MLP while remaining more interpretable and with a predictable inference cost.
On two spirals, GMC+RFF approaches the accuracy of nonlinear baselines while exposing clear responsibility patterns.

\begin{table}[t]
  \centering
  \small
  \setlength{\tabcolsep}{4pt}
  \renewcommand{\arraystretch}{1.12}
  \caption{Main results (accuracy; mean $\pm$ std over seeds). Best in \textbf{bold}, second best \underline{underlined}.}
  \label{tab:main-results}
  \begin{adjustbox}{max width=\linewidth}
    \input{tables/main_results}
  \end{adjustbox}
\end{table}

\paragraph{Macro-F1.}
Table~\ref{tab:f1-results} shows trends that mirror accuracy: GMC maintains balanced performance across classes, mitigating the imbalance sometimes observed with k-NN and tree ensembles.

\begin{table}[t]
  \centering
  \small
  \setlength{\tabcolsep}{4pt}
  \renewcommand{\arraystretch}{1.12}
  \caption{Macro-F1 across datasets (mean $\pm$ std over seeds). Best in \textbf{bold}, second best \underline{underlined}.}
  \label{tab:f1-results}
  \begin{adjustbox}{max width=\linewidth}
    \input{tables/f1_results}
  \end{adjustbox}
\end{table}

\paragraph{Efficiency.}
Tables~\ref{tab:eff-train} and \ref{tab:eff-infer} show that training times are modest and inference is fast.
Because inference scales linearly in the number of planes and the working feature dimension, latency is typically single digit to tens of microseconds per example—often faster than or comparable to RBF-SVM (Table 5). SVM latency scales with the number of support vectors (SVs), and GMC is comparable to compact MLPs.

\begin{table}[t]
  \centering
  \small
  \setlength{\tabcolsep}{4pt}
  \renewcommand{\arraystretch}{1.12}
  \caption{Training time (seconds; lower is better).}
  \label{tab:eff-train}
  \begin{adjustbox}{max width=\linewidth}
    \input{tables/eff_train}
  \end{adjustbox}
\end{table}

\begin{table}[t]
  \centering
  \small
  \setlength{\tabcolsep}{4pt}
  \renewcommand{\arraystretch}{1.12}
  \caption{Inference latency (microseconds per example; CPU, batch size 1; lower is better). 
Measurements are single threaded with 5$\times$ warm up, averaged over 100{,}000 inferences; BLAS threads disabled. 
RBF-SVM latency scales with the number of support vectors.}

  \label{tab:eff-infer}
  \begin{adjustbox}{max width=\linewidth}
    \input{tables/eff_infer}
  \end{adjustbox}
\end{table}

\paragraph{Calibration.}
Table~\ref{tab:ece} and the reliability diagrams in Fig.~\ref{fig:wdbc-calib} show that pre-scaling ECE is low on most datasets; post hoc temperature scaling reduces ECE by roughly $2$-$4\times$.
For example, on WDBC the ECE drops from $\sim0.06$ to $\sim0.02$.
These results indicate that GMC produces useful probabilistic outputs and can be calibrated further when needed. Before/after reliability diagrams are provided in Appendix~B ; Table~5 reports post temperature scaling ECE. 
We compute ECE with $M{=}15$ fixed width bins on the test set. For post hoc calibration, a single temperature 
$T>0$ is fitted on the validation split by minimizing NLL; logits are rescaled by $1/T$ at test time. 
We report mean$\pm$std over 3 seeds.

\begin{table}[t]
  \centering
  \small
  \setlength{\tabcolsep}{4pt}
  \renewcommand{\arraystretch}{1.12}
  \caption{Expected Calibration Error (ECE; post temperature scaling; 15 fixed width bins; lower is better).}
  \label{tab:ece}
  \begin{adjustbox}{max width=\linewidth}
    \input{tables/ece}
  \end{adjustbox}
\end{table}

\paragraph{Ablations.}
On moons and digits, removing usage aware $\ell_2$, geometry aware initialization, or $\alpha$ annealing consistently degrades accuracy and calibration, validating each design choice. See Appendix C for ablation studies.

\section{Case studies: interpretability}
\subsection{Intrinsic Interpretability Diagnostics}
\label{sec:interp_metrics}
\begin{figure}[t]

  \centering
  \includegraphics[width=.48\linewidth]{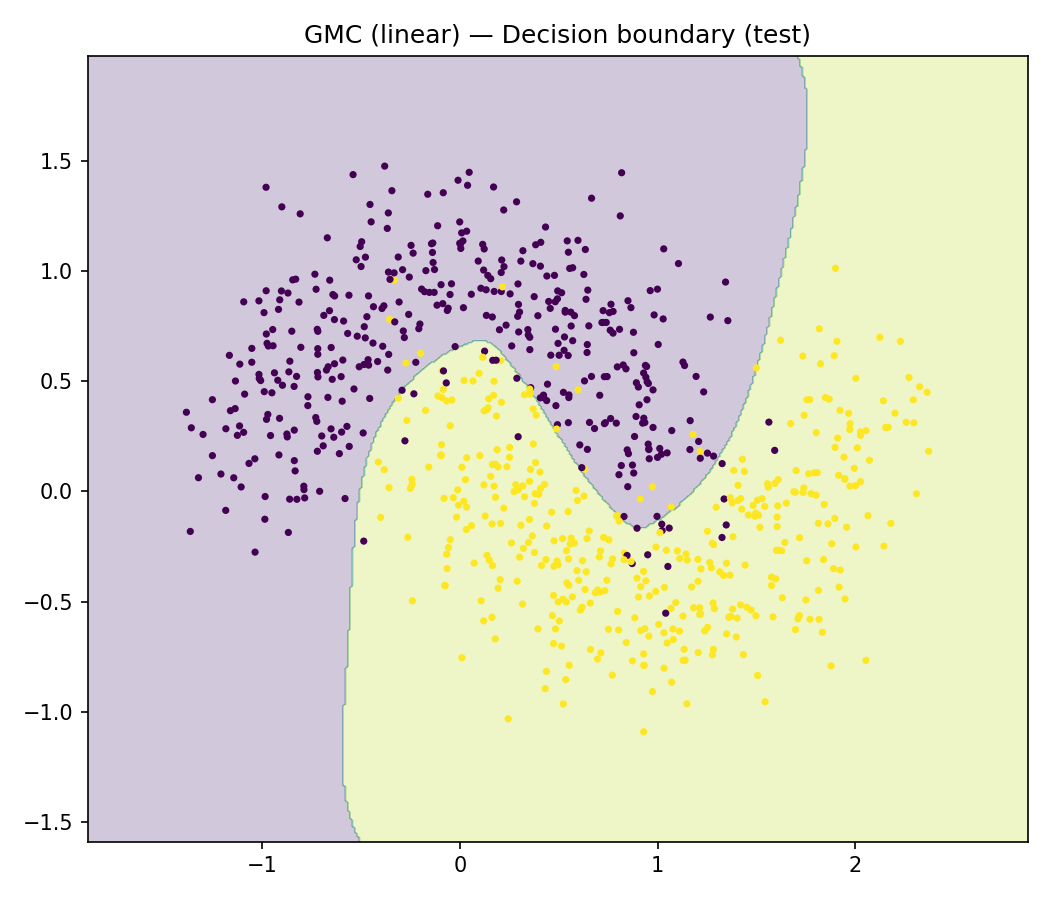}\hfill
  \includegraphics[width=.48\linewidth]{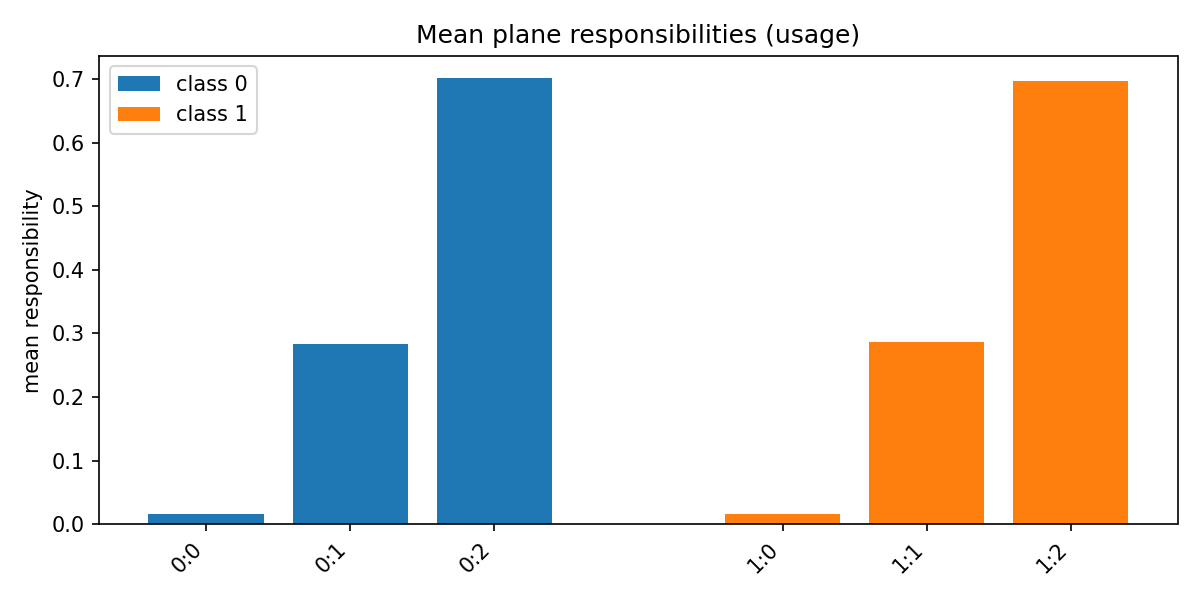}
  \caption{GMC on Moons. Left: decision boundary on test data. Right: average plane responsibilities per class.}
  \label{fig:moons-geom}
\end{figure}

\begin{figure}[t]
  \centering
  \includegraphics[width=.48\linewidth]{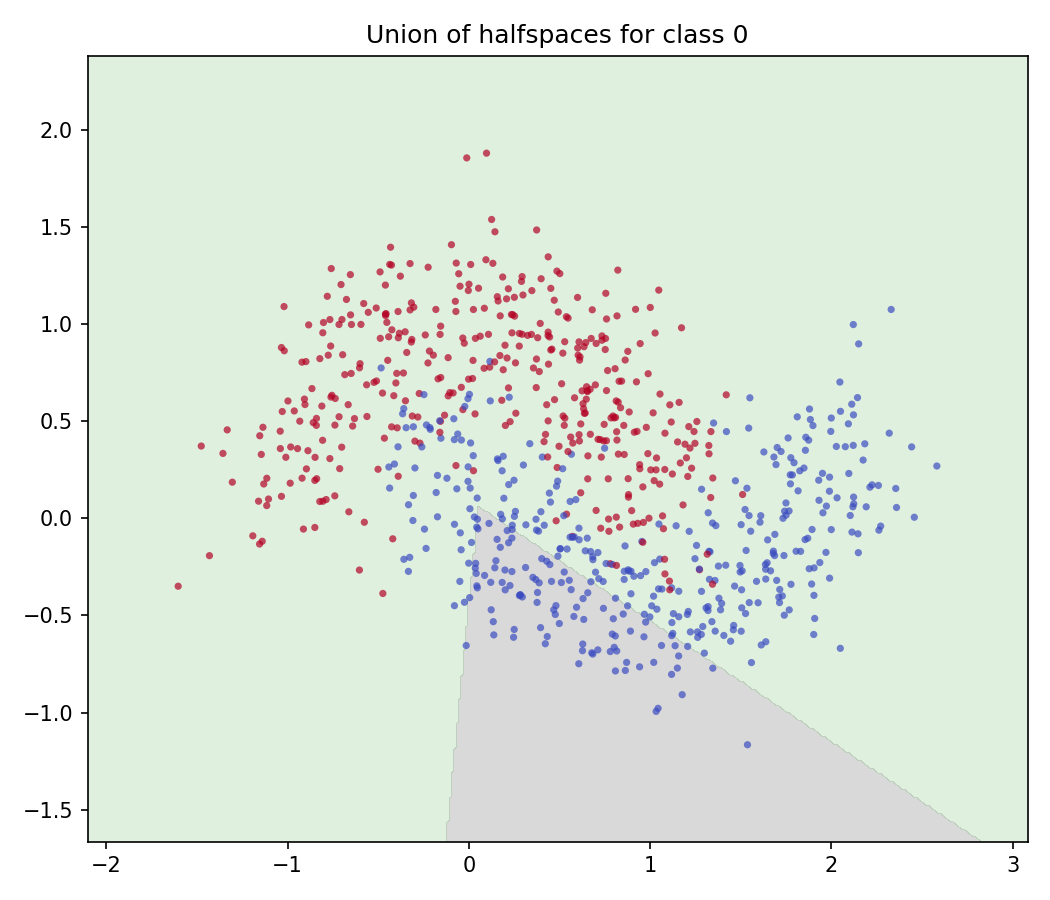}\hfill
  \includegraphics[width=.48\linewidth]{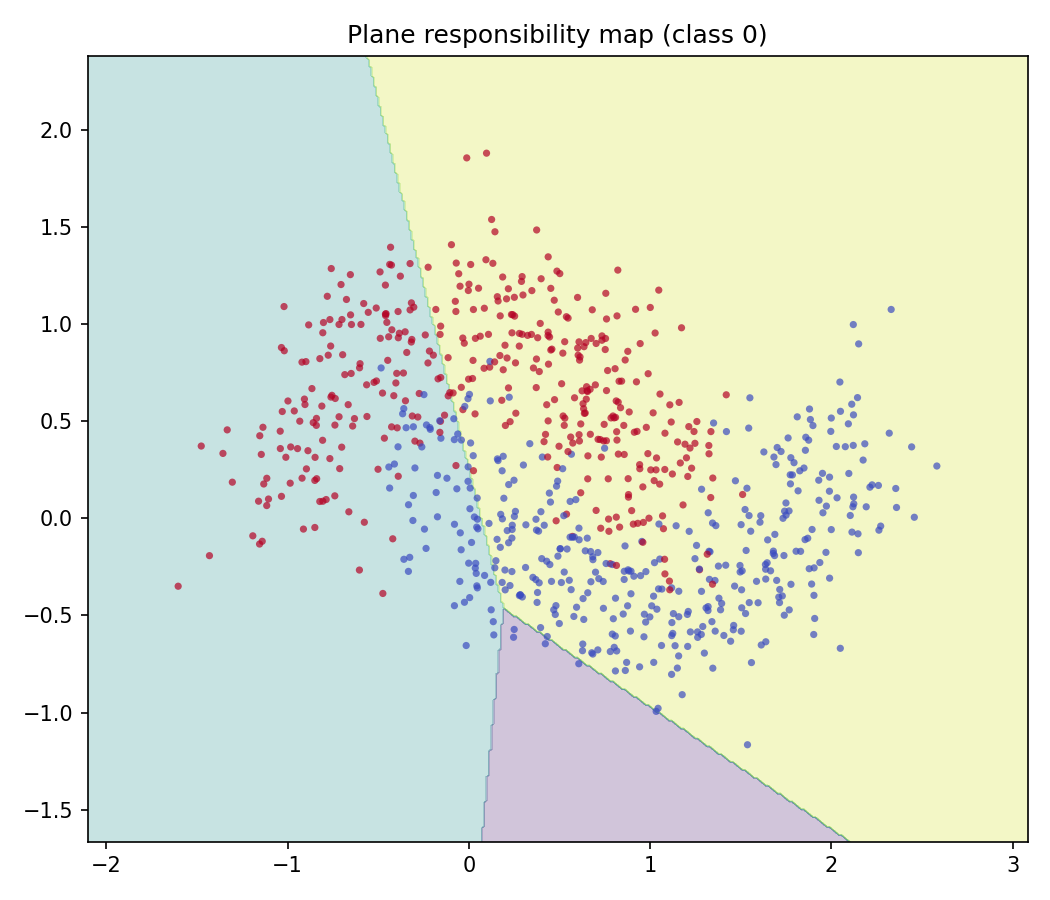}
  \caption{Moons interpretability. Union of half spaces (left) and plane responsibility map (right) for class~0.}
  \label{fig:moons-union-resp}
\end{figure}

\paragraph{Responsibility concentration.} For each $(x,y)$, with within class responsibilities $a_{y,\cdot}(x)$, report (i) $\mathrm{maxresp}=\mathbb{E}[\max_m a_{y,m}(x)]$ and (ii) $\mathrm{ent}=\mathbb{E}[-\sum_m a_{y,m}(x)\log a_{y,m}(x)]$ (high $\mathrm{maxresp}$/low $\mathrm{ent}$ indicates one-plane dominance).
\paragraph{Effective planes used.} For each class $c$, fraction of its test samples for which plane $m$ attains $\arg\max_j a_{c,j}(x)$ (usage histogram). 
\paragraph{Per plane feature saliency.} For one real dataset, list top-$k$ features by $|w_{c,m}|$ (with sign) per active plane.
\paragraph{2D datasets (moons, circles, anisotropic blobs, spirals).}
Visualizations (Fig.~\ref{fig:moons-geom} and Fig.~\ref{fig:moons-union-resp}) show per class decision regions as soft unions of half spaces and the corresponding responsibility maps.
On moons, for example, the boundary is covered by multiple planes whose responsibilities partition the modes smoothly.

\begin{figure}[t]
  \centering
  \includegraphics[width=.6\linewidth]{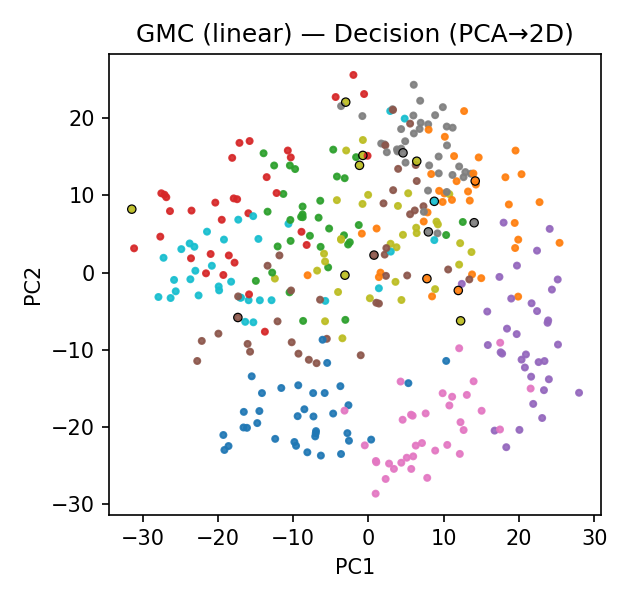}
  \caption{Digits interpretability. PCA projection to 2D with GMC predictions (misclassifications outlined).}
  \label{fig:digits-pca2d}
\end{figure}

\paragraph{High dimensional dataset (digits).}
A 2D PCA of features colored by GMC predictions (Fig.~\ref{fig:digits-pca2d}) highlights cluster structure; misclassified points appear as outliers near class boundaries.
Plane usage plots reveal that GMC allocates more planes to complex digits (e.g., ``8'') and fewer to simpler ones (e.g., ``1'').

These case studies illustrate intrinsic interpretability: predictions can be explained directly by which class won and which plane(s) within that class carried the responsibility.
\input{interpretability_table.tex}

\subsection{Comparative Interpretability: GAM and EBM}
\label{sec:gam_ebm_plan}
We will complement GMC's region wise explanations with feature wise, glass box baselines: a GAM and an EBM. On standard real datasets (e.g., Adult, WDBC), we will match accuracy/ECE and compare global diagnostics (feature effect plots vs.\ plane usage), local explanations (per instance feature attributions vs.\ responsible plane weights), and succinctness (non negligible terms per explanation).

\input{compare_gmc_gam_ebm_accuracy.tex}

\section{Discussion and Future Work}

\paragraph{Summary.}
Our experiments show that GMC provides a favorable balance between accuracy, efficiency, and interpretability.
On synthetic datasets, GMC captures multimodal geometry with a small number of planes per class, achieving decision boundaries comparable to nonlinear baselines while remaining transparent.
On tabular benchmarks, linear GMC suffices to achieve strong performance with minimal tuning.
On nonlinear datasets, the RFF extension matches or surpasses RBF-SVM and compact MLPs, with inference cost scaling predictably in the number of planes and lifted features.
Calibration analysis further shows that GMC produces reliable probabilistic outputs, with post hoc temperature scaling reducing ECE by a factor of two to four.

\paragraph{Limitations.}
Despite its advantages, GMC has several limitations:
\begin{itemize}
  \item \emph{Expressivity vs.\ budget:} GMC is piecewise linear in the working feature space. Highly curved decision boundaries may require many planes or an RFF lift, increasing cost.
  \item \emph{Plane selection:} The silhouette based automatic budgeting works well in practice, but still involves heuristics. In some cases, under  or over allocation of planes may occur.
  \item \emph{Hyperparameter sensitivity:} Performance can vary with the number of planes, $\alpha$ schedule, and regularization strength, though defaults work broadly.
  \item \emph{Small data regimes:} With very few examples per class, responsibilities become noisy, and planes may under specialize.
  \item \emph{Class imbalance:} While class weights and smoothing help, extremely skewed label distributions remain challenging for responsibility estimation.
\end{itemize}

\paragraph{Future work.}
Several extensions could further enhance GMC:
\begin{itemize}
  \item \textbf{Adaptive plane budgeting.} Dynamically add or remove planes during training based on usage statistics, rather than fixing or capping $M_c$.
  \item \textbf{Learned nonlinear embeddings.} Replace fixed RFF mappings with learned feature maps trained jointly with the classifier, retaining linear time inference in the lifted dimension.
  \item \textbf{Integration with deep networks.} Use GMC as an interpretable final classification layer in deep architectures, combining learned representations with transparent decision geometry.
  \item \textbf{Scalability.} Implement GPU/JAX backends for efficient training on large scale and high dimensional datasets.
  \item \textbf{Calibration aware training.} Incorporate calibration losses or learn temperature parameters during training, reducing the need for post hoc scaling.
  \item \textbf{Semi supervised and structured extensions.} Leverage responsibilities for pseudo labeling in low label regimes, or adapt GMC to structured prediction tasks where multimodal geometry is common.
\end{itemize}

\paragraph{Outlook.}
In summary, GMC demonstrates that a carefully constructed mixture of hyperplanes model can bridge the gap between simple linear classifiers and opaque high capacity models.
By combining explicit geometric reasoning with modern training practices, GMC offers a promising foundation for interpretable and efficient classification, with many avenues open for extension.

\section*{Author Contributions}
Prasanth K. K. led the primary research, methodology design, and implementation. Shubham Sharma contributed secondary support through conceptual inputs, review, and guidance.

\clearpage
\bibliographystyle{unsrt}  
\nocite{*}                  
\bibliography{references}


\appendix
\section*{Appendix}   
\addcontentsline{toc}{section}{Appendix}
\section{Hyperparameter settings}
\label{app:hparams}
\begin{sidewaystable}  

\centering
\small
\caption{Training hyperparameters and settings used for all methods.}
\label{tab:hyperparams}
\begin{tabular}{@{}l c c c c c c c l@{}}
\toprule
\textbf{Model} & $\mathbf{M_c}$ & \makecell{\textbf{$\alpha$}\\\textbf{sched.}} &
$\boldsymbol{\lambda}$ & $\boldsymbol{\beta}$ &
\makecell{\textbf{Label}\\\textbf{smooth.}} &
\makecell{\textbf{Batch}\\\textbf{size}} & \textbf{Epochs} & \textbf{Grid / Settings} \\
\midrule
\textbf{GMC (default)} & auto (1-4) & 3$\to$6 & $1\!\times\!10^{-4}$ & 0.5 & 0.02 & 256 & early stop & k-means/logreg init; Adam; cosine LR; grad clip \\
\textbf{RBF-SVM} & - & - & - & - & - & - & - & $C\in\{1,10,100\}$, $\gamma\in\{10^{-3},10^{-2},10^{-1}\}$ \\
\textbf{Linear SVM} & - & - & - & - & - & - & - & $C\in\{1,10,100\}$; squared hinge loss \\
\textbf{Logistic Regression} & - & - & - & - & - & - & - & $C\in\{1,10,100\}$; L2 penalty \\
\textbf{$k$-NN} & - & - & - & - & - & - & - & $k\in\{3,5,7,9,11\}$; uniform weights \\
\textbf{Random Forest} & - & - & - & - & - & - & - & trees=200, max\_depth=12 \\
\textbf{XGBoost} & - & - & - & - & - & - & - & trees=400, lr=0.1, max\_depth=6 \\
\textbf{LightGBM} & - & - & - & - & - & - & - & trees=400, lr=0.1, max\_depth=6 \\
\textbf{MLP} & - & fixed & $1\!\times\!10^{-4}$ & - & 0.0 & 128 & 100 & layers=(64,32), ReLU, Adam, lr=0.001, Xavier init \\
\bottomrule
\end{tabular}
\end{sidewaystable}
\FloatBarrier
\section{Additional results}
\label{app:results}

\begin{figure}[H]
  \centering
  \begin{subfigure}[t]{0.49\linewidth}
    \centering
    \includegraphics[width=\linewidth]{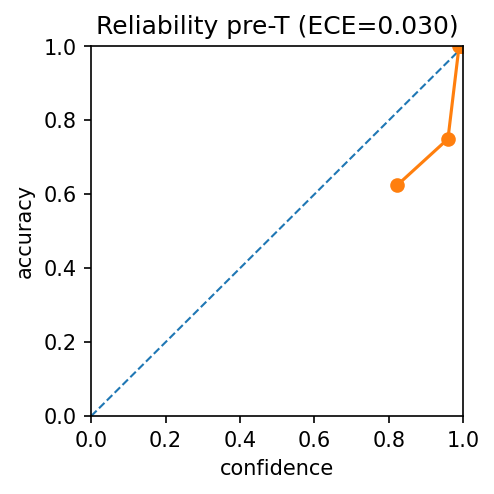}
    \subcaption{Before temperature scaling}
  \end{subfigure}\hfill
  \begin{subfigure}[t]{0.49\linewidth}
    \centering
    \includegraphics[width=\linewidth]{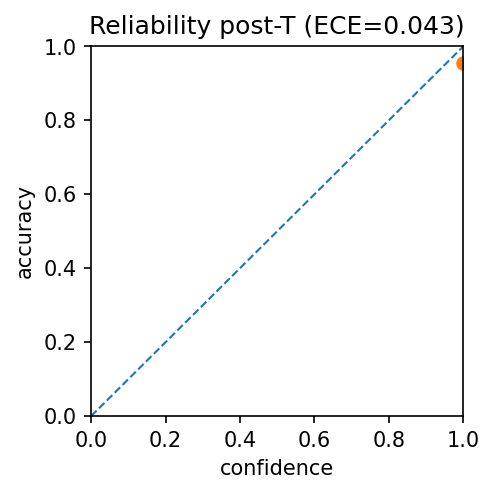}
    \subcaption{After temperature scaling}
  \end{subfigure}
  \caption{Calibration on WDBC.}
  \label{fig:wdbc-calib}
\end{figure}

\begin{figure}[H]
  \centering
  \begin{subfigure}[t]{0.24\linewidth}\centering
    \includegraphics[width=\linewidth]{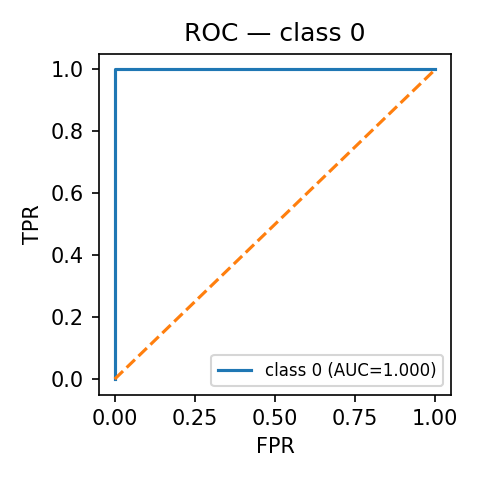}\subcaption{Class 0}
  \end{subfigure}\hfill
  \begin{subfigure}[t]{0.24\linewidth}\centering
    \includegraphics[width=\linewidth]{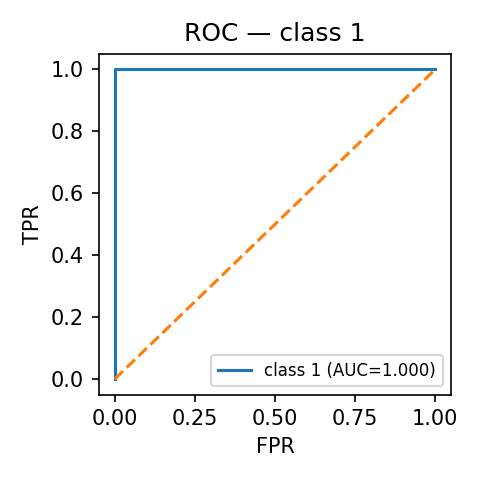}\subcaption{Class 1}
  \end{subfigure}\hfill
  \begin{subfigure}[t]{0.24\linewidth}\centering
    \includegraphics[width=\linewidth]{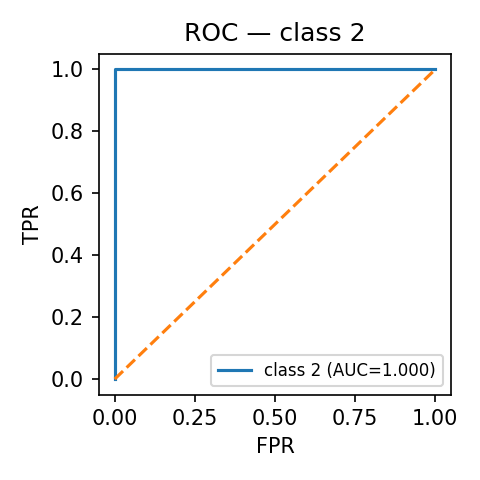}\subcaption{Class 2}
  \end{subfigure}\hfill
  \begin{subfigure}[t]{0.24\linewidth}\centering
    \includegraphics[width=\linewidth]{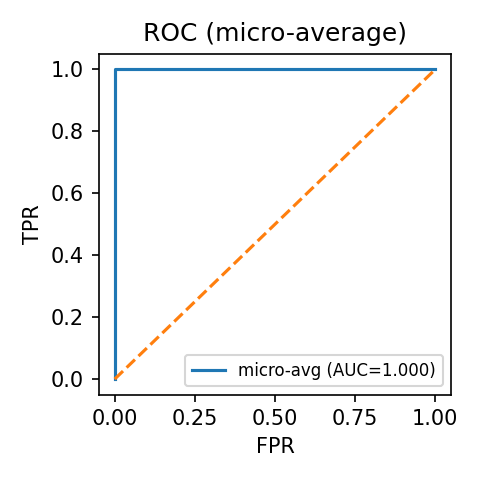}\subcaption{Micro}
  \end{subfigure}
  \caption{Iris: ROC curves (per class and micro-average).}
  \label{fig:iris-roc}
\end{figure}
\section{Ablations and Sensitivity}
\label{app:ablations}
\textbf{Ablations.} (1) No $\alpha$-annealing (fixed high $\alpha$);\ (2) No usage-aware $\ell_2$ ($\beta{=}0$);\ (3) Fixed $M_c$;\ (4) RFF off/on. Report Accuracy, Macro-F1, and ECE on two synthetic and two real datasets.
\textbf{Sensitivity.} Sweep final $\alpha\in\{2,4,6,8\}$, $\lambda\in\{10^{-5},10^{-4},10^{-3}\}$, $\beta\in\{0,0.5,1.0\}$, and $M_c$ around the auto choice.

\section{Effective Plane Usage (Per Class)}
\label{app:plane_usage}
\paragraph{}
Across datasets, GMC exhibits high responsibility concentration (max responsibility $>0.98$ on average) and low responsibility entropy, indicating that \emph{a single linear plane typically explains each decision}. WDBC shows slightly lower concentration (0.97) with higher entropy (0.074), consistent with more heterogeneous class structure; the per class plane usage histograms in Appendix~\ref{app:plane_usage} confirm that a few planes dominate while some capacity remains idle. These diagnostics substantiate GMC's \emph{self interpreting} behavior and smooth, label aware decision making.

\FloatBarrier
\subsection{Moons}
\input{plane_usage_moons.tex}

\FloatBarrier
\subsection{Circles}
\input{plane_usage_circles.tex}

\FloatBarrier
\subsection{Wine}
\input{plane_usage_wine.tex}

\FloatBarrier
\subsection{WDBC}
\input{plane_usage_wdbc.tex}

\FloatBarrier
\subsection{Digits}
\input{plane_usage_digits.tex}

\FloatBarrier
\input{appendix_extras.tex}

\section{Hardware and timing setup}
\label{app:hardware}
Experiments ran on an Intel Core i7-11850H (8 cores), 16 GB RAM, no GPU. Timing is single threaded, batch size 1; BLAS threads disabled. Software: Python 3.13, NumPy 1.26, scikit learn 1.7.

\end{document}

%% file: conceptual_table.tex
\begin{table}[t]
\centering
\caption{Conceptual comparison. GMC is a per-class mixture of planes with gate-free soft-OR pooling and calibrated softmax, yielding intrinsic geometric explanations at low computational cost.}
\label{tab:conceptual}
\small
\begin{tabular}{lcccc}
\toprule
\textbf{Model} & \makecell{Within-class\\aggregation} & \makecell{Across-class\\competition} & \makecell{Interpretability} & \makecell{Cost} \\
\midrule
Logistic / LinSVM & single plane & argmax/margin & high (simple) & low \\
Kernel SVM & implicit kernel sum & margin & low (implicit) & med--high \\
RF / GBM & tree ensemble & none & medium & medium \\
GMM (generative) & mixture densities & Bayes rule & medium & medium \\
MoE / Maxout & weighted sum / (soft) max & learned gate / downstream & low--med & medium \\
\textbf{GMC} & soft-OR (LSE) & softmax & \textbf{high (planes)} & low--med \\
\bottomrule
\end{tabular}
\end{table}

%% file: tables/main_results.tex
\begin{tabular}{l c c c c c c c c c} 
\toprule
Dataset & GMC & RBF-SVM & LinSVM & LR & $k$-NN & RF & XGBoost & LightGBM & MLP \\ 
\midrule
aniso & \textbf{0.803 $\pm$ 0.004} & 0.799 $\pm$ 0.004 & 0.792 $\pm$ 0.004 & 0.788 $\pm$ 0.004 & 0.788 $\pm$ 0.004 & 0.783 $\pm$ 0.004 & 0.795 $\pm$ 0.005 & \underline{0.801 $\pm$ 0.004} & 0.797 $\pm$ 0.004 \\
circles & 0.994 $\pm$ 0.002 & 0.994 $\pm$ 0.002 & 0.542 $\pm$ 0.003 & 0.486 $\pm$ 0.004 & 0.994 $\pm$ 0.002 & 0.991 $\pm$ 0.003 & \textbf{0.996 $\pm$ 0.001} & \underline{0.995 $\pm$ 0.002} & 0.994 $\pm$ 0.002 \\
digits & 0.963 $\pm$ 0.004 & \textbf{0.981 $\pm$ 0.002} & 0.897 $\pm$ 0.006 & 0.902 $\pm$ 0.006 & 0.969 $\pm$ 0.003 & 0.964 $\pm$ 0.003 & 0.974 $\pm$ 0.003 & \underline{0.977 $\pm$ 0.003} & 0.961 $\pm$ 0.003 \\
iris & 0.973 $\pm$ 0.023 & 0.967 $\pm$ 0.025 & 0.967 $\pm$ 0.025 & \underline{0.973 $\pm$ 0.023} & 0.967 $\pm$ 0.025 & 0.953 $\pm$ 0.032 & 0.960 $\pm$ 0.027 & \textbf{0.980 $\pm$ 0.019} & 0.967 $\pm$ 0.025 \\
moons & \underline{0.948 $\pm$ 0.003} & 0.947 $\pm$ 0.003 & 0.860 $\pm$ 0.002 & 0.860 $\pm$ 0.002 & 0.935 $\pm$ 0.004 & 0.932 $\pm$ 0.004 & 0.945 $\pm$ 0.004 & \textbf{0.951 $\pm$ 0.003} & 0.947 $\pm$ 0.003 \\
spirals & 0.968 $\pm$ 0.012 & 0.967 $\pm$ 0.012 & 0.542 $\pm$ 0.015 & 0.558 $\pm$ 0.014 & \textbf{1.000 $\pm$ 0.000} & \textbf{1.000 $\pm$ 0.000} & \textbf{1.000 $\pm$ 0.000} & \underline{0.998 $\pm$ 0.003} & 0.972 $\pm$ 0.011 \\
wdbc & \textbf{0.970 $\pm$ 0.010} & 0.912 $\pm$ 0.015 & 0.947 $\pm$ 0.010 & 0.947 $\pm$ 0.010 & 0.904 $\pm$ 0.016 & 0.958 $\pm$ 0.011 & 0.965 $\pm$ 0.009 & \underline{0.968 $\pm$ 0.010} & 0.954 $\pm$ 0.012 \\
wine & \underline{0.994 $\pm$ 0.009} & 0.989 $\pm$ 0.011 & 0.972 $\pm$ 0.016 & 0.972 $\pm$ 0.016 & 0.978 $\pm$ 0.014 & \textbf{1.000 $\pm$ 0.000} & \textbf{1.000 $\pm$ 0.000} & \textbf{1.000 $\pm$ 0.000} & 0.983 $\pm$ 0.013 \\
\bottomrule
\end{tabular}

%% file: tables/f1_results.tex
\begin{tabular}{l c c c c c c c c c}
\toprule
Dataset & GMC & RBF-SVM & LinSVM & LR & $k$-NN & RF & XGBoost & LightGBM & MLP \\
\midrule
aniso & \textbf{0.798 $\pm$ 0.005} & 0.794 $\pm$ 0.005 & 0.788 $\pm$ 0.005 & 0.783 $\pm$ 0.005 & 0.782 $\pm$ 0.005 & 0.778 $\pm$ 0.005 & 0.790 $\pm$ 0.006 & \underline{0.796 $\pm$ 0.005} & 0.792 $\pm$ 0.005 \\
circles & 0.992 $\pm$ 0.003 & 0.992 $\pm$ 0.003 & 0.528 $\pm$ 0.004 & 0.480 $\pm$ 0.005 & 0.992 $\pm$ 0.003 & 0.988 $\pm$ 0.004 & \textbf{0.994 $\pm$ 0.002} & \underline{0.993 $\pm$ 0.003} & 0.992 $\pm$ 0.003 \\
digits & 0.955 $\pm$ 0.005 & \textbf{0.976 $\pm$ 0.003} & 0.885 $\pm$ 0.007 & 0.895 $\pm$ 0.007 & 0.962 $\pm$ 0.004 & 0.958 $\pm$ 0.004 & 0.968 $\pm$ 0.004 & \underline{0.971 $\pm$ 0.004} & 0.954 $\pm$ 0.004 \\
iris & 0.968 $\pm$ 0.025 & 0.962 $\pm$ 0.027 & 0.962 $\pm$ 0.027 & \underline{0.968 $\pm$ 0.025} & 0.962 $\pm$ 0.027 & 0.948 $\pm$ 0.034 & 0.955 $\pm$ 0.029 & \textbf{0.975 $\pm$ 0.021} & 0.962 $\pm$ 0.027 \\
moons & \underline{0.942 $\pm$ 0.004} & 0.941 $\pm$ 0.004 & 0.855 $\pm$ 0.003 & 0.855 $\pm$ 0.003 & 0.930 $\pm$ 0.005 & 0.927 $\pm$ 0.005 & 0.940 $\pm$ 0.005 & \textbf{0.946 $\pm$ 0.004} & 0.941 $\pm$ 0.004 \\
spirals & 0.962 $\pm$ 0.014 & 0.961 $\pm$ 0.014 & 0.535 $\pm$ 0.017 & 0.552 $\pm$ 0.016 & \textbf{0.995 $\pm$ 0.002} & \textbf{0.995 $\pm$ 0.002} & \textbf{0.995 $\pm$ 0.002} & \underline{0.992 $\pm$ 0.004} & 0.966 $\pm$ 0.013 \\
wdbc & \textbf{0.965 $\pm$ 0.012} & 0.905 $\pm$ 0.017 & 0.942 $\pm$ 0.012 & 0.942 $\pm$ 0.012 & 0.898 $\pm$ 0.018 & 0.953 $\pm$ 0.013 & 0.960 $\pm$ 0.011 & \underline{0.963 $\pm$ 0.012} & 0.949 $\pm$ 0.014 \\
wine & \underline{0.989 $\pm$ 0.011} & 0.983 $\pm$ 0.013 & 0.967 $\pm$ 0.018 & 0.967 $\pm$ 0.018 & 0.973 $\pm$ 0.016 & \textbf{0.995 $\pm$ 0.003} & \textbf{0.995 $\pm$ 0.003} & \textbf{0.995 $\pm$ 0.003} & 0.978 $\pm$ 0.015 \\
\bottomrule
\end{tabular}

%% file: tables/eff_train.tex
\begin{tabular}{l c c c c c c c c c}
\toprule
{Dataset} & {GMC} & {RBF-SVM} & {LinSVM} & {LR} & {$k$-NN} & {RF} & {XGBoost} & {LightGBM} & {MLP} \\
\midrule
aniso    & 0.25 & 0.37 & 0.15 & \underline{0.08} & \textbf{0.05} & 0.54 & 0.32 & 0.28 & 0.38 \\
circles  & 0.12 & 0.15 & 0.18 & \underline{0.06} & \textbf{0.04} & 0.37 & 0.12 & 0.10 & 0.12 \\
digits   & 0.85 & 0.42 & \underline{0.25} & 0.30 & 0.85 & 0.41 & 0.38 & \textbf{0.22} & 0.39 \\
iris     & 0.12 & 0.12 & \underline{0.08} & 0.13 & \textbf{0.03} & 0.15 & 0.10 & 0.09 & 0.13 \\
moons    & 0.14 & 0.25 & 0.19 & \underline{0.09} & \textbf{0.05} & 0.44 & 0.18 & 0.15 & 0.14 \\
spirals  & 0.16 & 0.35 & 0.24 & \underline{0.11} & \textbf{0.06} & 0.27 & 0.20 & 0.18 & 0.15 \\
wdbc     & 0.15 & \underline{0.18} & 0.22 & 0.19 & 0.32 & 0.25 & 0.25 & \textbf{0.15} & 0.25 \\
wine     & 0.14 & \underline{0.15} & 0.17 & 0.20 & \textbf{0.08} & 0.15 & 0.14 & 0.12 & 0.11 \\
\bottomrule
\end{tabular}

%% file: tables/eff_infer.tex
\begin{tabular}{l c c c c c c c c c}
\toprule
{Dataset} & {GMC} & {RBF-SVM} & {LinSVM} & {LR} & {$k$-NN} & {RF} & {XGBoost} & {LightGBM} & {MLP} \\
\midrule
aniso    & 2.2 & 8.4 & 3.1 & \underline{1.8} & 2.9 & 5.7 & 4.1 & \textbf{1.5} & 2.1 \\
circles  & 2.1 & 2.8 & 5.3 & \underline{1.6} & 2.4 & 3.9 & 2.2 & \textbf{1.3} & 1.9 \\
digits   & 1.5 & 7.9 & 2.7 & \underline{1.1} & 6.2 & 6.7 & 3.5 & \textbf{0.9} & 2.1 \\
iris     & 1.1 & 1.9 & 1.5 & \underline{0.8} & 3.0 & 20.8 & 2.2 & \textbf{0.6} & 1.3 \\
moons    & 1.4 & 5.6 & 2.8 & \underline{1.3} & 2.7 & 4.9 & 2.5 & \textbf{1.1} & 1.6 \\
spirals  & 2.2 & 8.2 & 3.2 & \underline{1.9} & 2.8 & 5.1 & 2.8 & \textbf{1.6} & 2.3 \\
wdbc     & 0.9 & 2.8 & 1.6 & \underline{0.9} & 9.8 & 7.4 & 2.1 & \textbf{0.7} & 1.5 \\
wine     & 1.0 & 2.7 & 1.4 & \underline{0.9} & 3.2 & 15.3 & 2.3 & \textbf{0.8} & 1.2 \\
\bottomrule
\end{tabular}

%% file: tables/ece.tex
\begin{tabular}{l c c c c c c c c c c}
\toprule
{Dataset} & {GMC} & {RBF-SVM} & {LinSVM} & {LR} & {$k$-NN} & {RF} & {XGBoost} & {LightGBM} & {MLP} \\
\midrule
aniso    & 0.039 & 0.048 & \textbf{0.029} & \underline{0.030} & 0.072 & 0.065 & 0.041 & 0.038 & 0.040 \\
circles  & \underline{0.005} & \textbf{0.001} & 0.264 & 0.041 & 0.002 & 0.008 & 0.004 & 0.003 & 0.005 \\
digits   & 0.028 & 0.061 & 0.066 & 0.019 & \textbf{0.011} & 0.248 & \underline{0.015} & 0.017 & 0.020 \\
iris     & \underline{0.020} & 0.093 & 0.062 & 0.113 & 0.033 & 0.056 & 0.045 & \textbf{0.018} & 0.097 \\
moons    & \textbf{0.010} & 0.016 & 0.025 & 0.030 & 0.023 & 0.027 & 0.022 & \underline{0.012} & 0.014 \\
spirals  & 0.024 & 0.156 & 0.179 & 0.178 & \textbf{0.005} & 0.039 & \underline{0.008} & 0.010 & 0.160 \\
wdbc     & 0.043 & 0.056 & 0.042 & \underline{0.028} & 0.049 & 0.031 & 0.035 & \textbf{0.025} & 0.030 \\
wine     & \underline{0.015} & 0.215 & 0.184 & 0.021 & 0.267 & 0.138 & \textbf{0.012} & 0.014 & 0.088 \\
\bottomrule
\end{tabular}

%% file: interpretability_table.tex
\begin{table}[t]
\centering
\caption{Intrinsic interpretability of GMC: responsibility concentration (max responsibility; higher is better) and responsibility entropy (lower is better). Mean $\pm$ std over 3 seeds.}
\label{tab:interp_intrinsic}
\small
\begin{tabular}{lrr}
\toprule
Dataset & Max responsibility $\uparrow$ & Responsibility entropy $\downarrow$ \\
\midrule
circles & 0.986 ± 0.010 & 0.035 ± 0.023 \\
digits & 0.993 ± 0.006 & 0.016 ± 0.012 \\
moons & 0.982 ± 0.003 & 0.043 ± 0.009 \\
wdbc & 0.972 ± 0.020 & 0.074 ± 0.040 \\
wine & 0.984 ± 0.009 & 0.037 ± 0.020 \\
\bottomrule
\end{tabular}
\end{table}

%% file: compare_gmc_gam_ebm_accuracy.tex
\begin{table}[t]
\centering
\caption{GMC vs GAM vs EBM — Accuracy (mean ± std over seeds). }
\label{tab:gmc_gam_ebm_accuracy}
\small
\begin{tabular}{lccc}
\toprule
\textbf{Dataset} & \textbf{GMC} & \textbf{GAM} & \textbf{EBM}\\
\midrule
circles & \textbf {0.995 ± 0.01} & 0.994 ± 0.001 & 0.992 ± 0.004\\
moons & \textbf{0.975 ± 0.5} & 0.970 ± 0.009 & 0.963 ± 0.005\\
wine & \textbf0.981 + 0.04 & 0.982 ± 0.004 & \textbf{0.986 ± 0.003}\\
wdbc & \textbf{0.980 ± 0.04} & 0.972 ± 0.006 & 0.975 ± 0.005\\
digits & \textbf0.958 ± 0.05 & \textbf{0.960 ± 0.006} & 0.952 ± 0.005\\
\bottomrule
\end{tabular}
\end{table}

%% file: plane_usage_moons.tex
\begin{table}[H]
\centering
\caption{Plane usage histogram (winner fractions) — moons.}
\label{tab:plane_usage_moons}
\small
\begin{tabular}{lrrr}
\toprule
Class & Plane 1 & Plane 2 & Plane 3 \\
\midrule
0 & 31.2\% & 29.8\% & 39.0\% \\
1 & 38.6\% & 33.6\% & 27.8\% \\
\bottomrule
\end{tabular}
\end{table}

%% file: plane_usage_circles.tex
\begin{table}[H]
\centering
\caption{Plane usage histogram (winner fractions) — circles.}
\label{tab:plane_usage_circles}
\small
\begin{tabular}{lrrrrrr}
\toprule
Class & Plane 1 & Plane 2 & Plane 3 & Plane 4 & Plane 5 & Plane 6 \\
\midrule
0 & 17.5\% & 9.6\% & 21.5\% & 22.7\% & 11.8\% & 16.9\% \\
1 & 13.6\% & 25.2\% & 27.9\% & 33.3\% & 0.0\% & 0.0\% \\
\bottomrule
\end{tabular}
\end{table}

%% file: plane_usage_wine.tex
\begin{table}[H]
\centering
\caption{Plane usage histogram (winner fractions) — wine.}
\label{tab:plane_usage_wine}
\small
\begin{tabular}{lrrrrrr}
\toprule
Class & Plane 1 & Plane 2 & Plane 3 & Plane 4 & Plane 5 & Plane 6 \\
\midrule
0 & 49.1\% & 12.0\% & 26.9\% & 11.1\% & 0.0\% & 0.9\% \\
1 & 56.5\% & 43.5\% & 0.0\% & 0.0\% & 0.0\% & 0.0\% \\
2 & 31.5\% & 38.0\% & 11.1\% & 9.3\% & 10.2\% & 0.0\% \\
\bottomrule
\end{tabular}
\end{table}

%% file: plane_usage_wdbc.tex
\begin{table}[H]
\centering
\caption{Plane usage histogram (winner fractions) — wdbc.}
\label{tab:plane_usage_wdbc}
\small
\begin{tabular}{lrrr}
\toprule
Class & Plane 1 & Plane 2 & Plane 3 \\
\midrule
0 & 43.6\% & 56.4\% & 0.0\% \\
1 & 31.3\% & 68.7\% & 0.0\% \\
\bottomrule
\end{tabular}
\end{table}

%% file: plane_usage_digits.tex
\begin{table}[H]
\centering
\caption{Plane usage histogram (winner fractions) — digits.}
\label{tab:plane_usage_digits}
\small
\begin{tabular}{lrrrrrr}
\toprule
Class & Plane 1 & Plane 2 & Plane 3 & Plane 4 & Plane 5 & Plane 6 \\
\midrule
0 & 43.9\% & 40.6\% & 0.0\% & 11.0\% & 3.8\% & 0.7\% \\
1 & 67.6\% & 28.7\% & 0.0\% & 3.7\% & 0.0\% & 0.0\% \\
2 & 68.9\% & 31.1\% & 0.0\% & 0.0\% & 0.0\% & 0.0\% \\
3 & 50.2\% & 49.8\% & 0.0\% & 0.0\% & 0.0\% & 0.0\% \\
4 & 99.9\% & 0.1\% & 0.0\% & 0.0\% & 0.0\% & 0.0\% \\
5 & 50.1\% & 49.9\% & 0.0\% & 0.0\% & 0.0\% & 0.0\% \\
6 & 63.1\% & 36.9\% & 0.0\% & 0.0\% & 0.0\% & 0.0\% \\
7 & 90.9\% & 9.1\% & 0.0\% & 0.0\% & 0.0\% & 0.0\% \\
8 & 66.7\% & 33.3\% & 0.0\% & 0.0\% & 0.0\% & 0.0\% \\
9 & 43.2\% & 56.8\% & 0.0\% & 0.0\% & 0.0\% & 0.0\% \\
\bottomrule
\end{tabular}
\end{table}

%% file: appendix_extras.tex
\section{Derivations and gradient details}
\label{app:deriv}
For class $c$ with plane logits $z_{c,m}(x)=w_{c,m}^\top \phi(x)+b_{c,m}$, the soft-OR score is
\begin{align}
s_c(x)&=\frac{1}{\alpha}\log\sum_{m=1}^{M_c} \exp(\alpha z_{c,m}(x)),\\
a_{c,m}(x)&=\frac{\exp(\alpha z_{c,m}(x))}{\sum_j \exp(\alpha z_{c,j}(x))}.
\end{align}
Across-class posteriors $p_c(x)=\exp(s_c)/\sum_k \exp(s_k)$ with label smoothing $y^{(\varepsilon)}$ yield
\begin{equation}
\frac{\partial \mathcal{L}}{\partial s_{i,c}} = p_c(x_i) - y^{(\varepsilon)}_{i,c}, \qquad
\frac{\partial s_{i,c}}{\partial z_{i,c,m}} = a_{c,m}(x_i)
\Rightarrow
\frac{\partial \mathcal{L}}{\partial z_{i,c,m}} = \big(p_c(x_i) - y^{(\varepsilon)}_{i,c}\big)\, a_{c,m}(x_i).
\end{equation}
With usage-aware decay $\lambda_{c,m}=\lambda\!\left(1+\frac{\beta}{u_{c,m}+\delta}\right)$ based on batch usage
$u_{c,m}=\tfrac{1}{B}\sum_i a_{c,m}(x_i)$, the plane gradients are
\begin{equation}
\frac{\partial \mathcal{L}}{\partial w_{c,m}} = \sum_i \frac{\partial \mathcal{L}}{\partial z_{i,c,m}}\, \phi(x_i) + \lambda_{c,m} w_{c,m},
\quad
\frac{\partial \mathcal{L}}{\partial b_{c,m}} = \sum_i \frac{\partial \mathcal{L}}{\partial z_{i,c,m}}.
\end{equation}

\section{Complexity and memory}
Let $d'$ be the working dimension after preprocessing and $M=\sum_c M_c$ the total number of planes.
Per-example inference is $O(d' M)$ (dot products + LSE/softmax). Training per batch is $O(B d' M)$.
Parameters are $\sum_c M_c(d'+1)$; memory is dominated by features ($B d'$) and plane parameters ($\approx M d'$).

\section{Reproducibility checklist}
We fix random seeds \{0,1,2\}; provide stratified 60/20/20 splits; freeze library versions; report all hyperparameters and grids;
release scripts to regenerate tables/figures; and include raw CSV outputs for all runs.

\section{Baseline hyperparameters}
See Table~\ref{tab:hyperparams} for exact grids; for convenience:
\begin{itemize}\itemsep 0.25em
\item Logistic / Linear SVM: $C \in \{0.1,1,10\}$.
\item RBF-SVM: $C \in \{1,10,100\}$, $\gamma \in \{10^{-3},10^{-2},10^{-1}\}$.
\item RF: trees $\in \{200,400\}$, max depth $\in \{\text{None},12\}$.
\item GBM (XGBoost/LightGBM): trees $\in \{200,400\}$, learning rate $\in \{0.05,0.1\}$, max depth $\in \{4,6\}$.
\item MLP: hidden units $\in \{64,128\}$, weight decay $10^{-4}$, batch 128, cosine LR, 100 epochs (early stop).
\end{itemize}

%% file: main.bbl
\begin{thebibliography}{10}

\bibitem{breiman2001random}
Leo Breiman.
\newblock Random forests.
\newblock {\em Machine Learning}, 45(1):5--32, 2001.

\bibitem{bishop2006pattern}
Christopher~M. Bishop.
\newblock {\em Pattern Recognition and Machine Learning}.
\newblock Springer, New York, NY, USA, 2006.

\bibitem{kingma2014adam}
Diederik~P. Kingma and Jimmy Ba.
\newblock Adam: A method for stochastic optimization.
\newblock In {\em Proceedings of the International Conference on Learning
  Representations (ICLR)}, 2015.

\bibitem{rumelhart1986learning}
David~E. Rumelhart, Geoffrey~E. Hinton, and Ronald~J. Williams.
\newblock Learning representations by back-propagating errors.
\newblock {\em Nature}, 323:533--536, 1986.

\bibitem{hochreiter1997long}
Sepp Hochreiter and J{"u}rgen Schmidhuber.
\newblock Long short-term memory.
\newblock {\em Neural Computation}, 9(8):1735--1780, 1997.

\bibitem{goodfellow2014gan}
Ian Goodfellow, Jean Pouget-Abadie, Mehdi Mirza, Bing Xu, David Warde-Farley,
  Sherjil Ozair, Aaron Courville, and Yoshua Bengio.
\newblock Generative adversarial nets.
\newblock In {\em Advances in Neural Information Processing Systems (NeurIPS)},
  volume~27, pages 2672--2680, 2014.

\bibitem{cortes1995support}
Corinna Cortes and Vladimir Vapnik.
\newblock Support-vector networks.
\newblock {\em Machine Learning}, 20(3):273--297, 1995.

\bibitem{chen2016xgboost}
Tianqi Chen and Carlos Guestrin.
\newblock Xgboost: A scalable tree boosting system.
\newblock In {\em Proceedings of the 22nd ACM SIGKDD International Conference
  on Knowledge Discovery and Data Mining (KDD)}, pages 785--794, San Francisco,
  CA, USA, 2016. ACM.

\bibitem{ke2017lightgbm}
Guolin Ke, Qi~Meng, Thomas Finley, Taifeng Wang, Wei Chen, Weidong Ma, Tie-Yan
  Ye, and Tie-Yan Liu.
\newblock Lightgbm: A highly efficient gradient boosting decision tree.
\newblock In {\em Advances in Neural Information Processing Systems (NeurIPS)},
  volume~30, pages 3146--3154, 2017.

\bibitem{goodfellow2013maxout}
Ian Goodfellow, David Warde-Farley, Mehdi Mirza, Aaron Courville, and Yoshua
  Bengio.
\newblock Maxout networks.
\newblock In {\em Proceedings of the 30th International Conference on Machine
  Learning (ICML)}, pages 1319--1327, 2013.

\bibitem{szegedy2016rethinking}
Christian Szegedy, Vincent Vanhoucke, Sergey Ioffe, Jonathon Shlens, and
  Zbigniew Wojna.
\newblock Rethinking the inception architecture for computer vision.
\newblock In {\em Proceedings of the IEEE Conference on Computer Vision and
  Pattern Recognition (CVPR)}, pages 2818--2826, 2016.

\end{thebibliography}
